\documentclass{article}

\usepackage{arxiv}

\usepackage[utf8]{inputenc} % allow utf-8 input
\usepackage[T1]{fontenc}    % use 8-bit T1 fonts
\usepackage{hyperref}       % hyperlinks
\usepackage{url}            % simple URL typesetting
\usepackage{booktabs}       % professional-quality tables
\usepackage{amsfonts}       % blackboard math symbols
\usepackage{nicefrac}       % compact symbols for 1/2, etc.
\usepackage{microtype}      % microtypography
\usepackage{lipsum}		% Can be removed after putting your text content
\usepackage{graphicx}
\usepackage{natbib}
\usepackage{doi}
\usepackage{subcaption}
\usepackage{enumitem}

\title{Can virtual staining for high-throughput screening generalize?}

\author{ \href{https://orcid.org/0009-0008-1408-1493}{\includegraphics[scale=0.06]{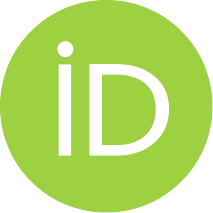}\hspace{1mm}Samuel Tonks} \\
	School of Computer Science,\\
	University of Birmingham,\\
	Birmingham, UK. \\
	\texttt{sxt118@student.bham.ac.uk} \\
	\And
	\href{https://orcid.org/0000-0002-5151-1893}{\includegraphics[scale=0.06]{orcid.pdf}\hspace{1mm} Cuong Nguyen} \\
	Artificial Intelligence \& Machine Learning, \\
        GSK, \\
        South San Francisco, \\
        California 94080, United States. \\
	\texttt{cuong.q.nguyen@gsk.com} \\
        \And
	\href{https://orcid.org/0000-0002-7708-7699}{\includegraphics[scale=0.06]{orcid.pdf}\hspace{1mm} Steve Hood} \\
	GSK Drug Metabolism \& Pharmacokinetics, \\ 
    GSK Medicines Research Centre, \\
        Gunnels Wood Road, \\
        Stevenage, Hertfordshire, SG1 2NY, UK. \\
	\texttt{steve.r.hood@gsk.com} \\
        \And
	{Ryan Musso} \\
	GSK Genome Biology, \\ 
        1250 S Collegeville Rd, \\
        Collegeville, PA 19426, United States. \\
	\texttt{ryan.x.musso@gsk.com} \\
        \And
	{Ceridwen Hopely} \\
	GSK Genome Biology, \\ 
        1250 S Collegeville Rd, \\
        Collegeville, PA 19426, United States. \\
	\texttt{ceridwen.s.hopely@gsk.com} \\
        \And
	{Steve Titus} \\
	GSK Genome Biology, \\ 
        1250 S Collegeville Rd, \\
        Collegeville, PA 19426, United States. \\
	\texttt{steve.titus@thermofisher.com} \\
        \And
	\href{https://orcid.org/0000-0002-3235-0457}{\includegraphics[scale=0.06]{orcid.pdf}\hspace{1mm} Minh Doan *} \\
	GSK Bioimaging, \\ 
        1250 S Collegeville Rd, \\
        Collegeville, PA 19426, United States. \\
	\texttt{minh.x.doan@gsk.com} \\
        \And
	\href{https://orcid.org/0000-0002-6755-0299}{\includegraphics[scale=0.06]{orcid.pdf}\hspace{1mm} Iain Styles *} \\
	School of Electronics, Electrical Engineering\\
 and Computer Science, \\
        Queen's University, \\
        Belfast, UK.\\
	\texttt{i.styles@qub.ac.uk} \\
        \And
	\href{https://orcid.org/0000-0002-7778-7169}
    {\includegraphics[scale=0.06]{orcid.pdf}\hspace{1mm} Alexander Krull *} \\
	School of Computer Science,\\
	University of Birmingham,\\
	Birmingham, UK. \\
	\texttt{a.f.f.krull@bham.ac.uk} \\
}

\renewcommand{\headeright}{}
\renewcommand{\shorttitle}{}

\newcommand\figtesaro{
\begin{figure*}[ht]
    \centering
    \vspace{-8mm}
    \includegraphics[width=1\linewidth]{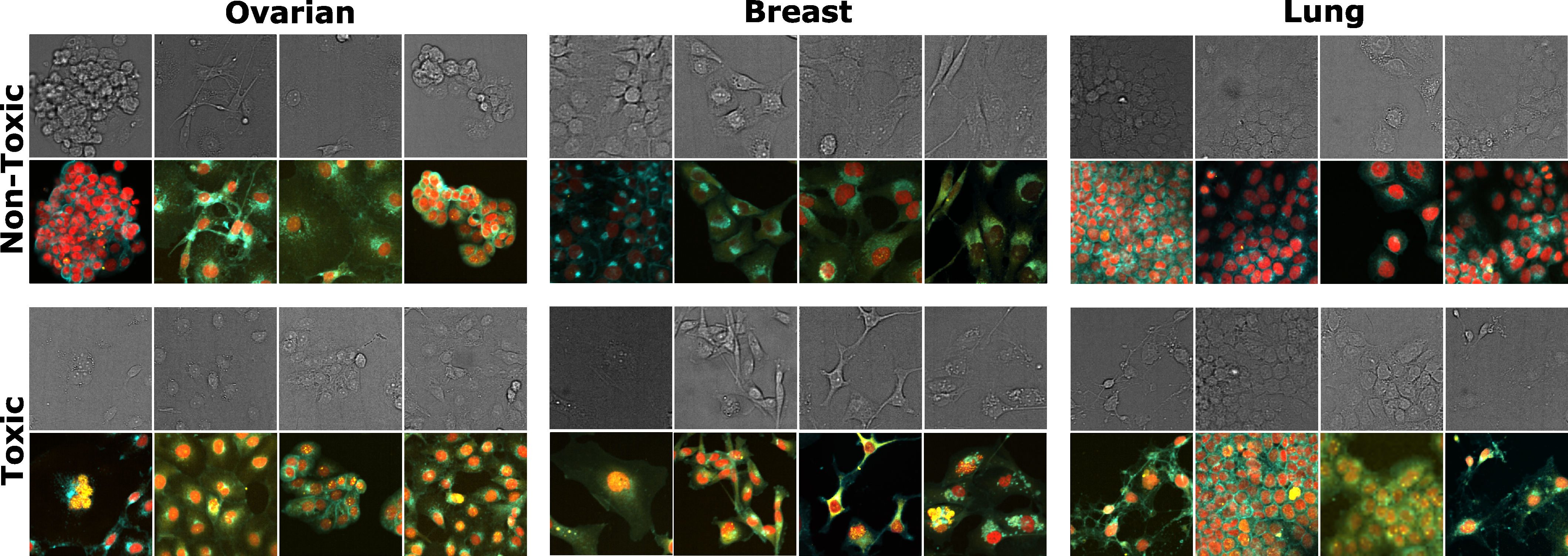}
    \caption{\textbf{GSK HTS dataset comprised of three different cell types; ovarian, breast, and lung and two phenotypes; non-toxic and toxic}. Each 2x4 is comprised of 4 randomly selected bright-field and fluorescence stain image pairs (shown as a composite image) for each cell type and phenotype. The composite image shows the nuclei stain (DAPI) in red, the cytoplasm stain (FITC) in cyan, and the DNA-damage stain (Cy5) in yellow. Within the dataset, we observe variability within the toxic and non-toxic samples of each cell type as well as anatomical differences between the different cell types. We explore the generalization performance across all three virtual staining tasks for three common HTS data distribution shifts; generalizing to new phenotypes, generalizing to new cell types, and both combined.}
    \label{fig:dataset}
\end{figure*}
}

\newcommand\figPhenotypevisuals{
\begin{figure*}[ht]
    \centering
    \includegraphics[width=1\linewidth]{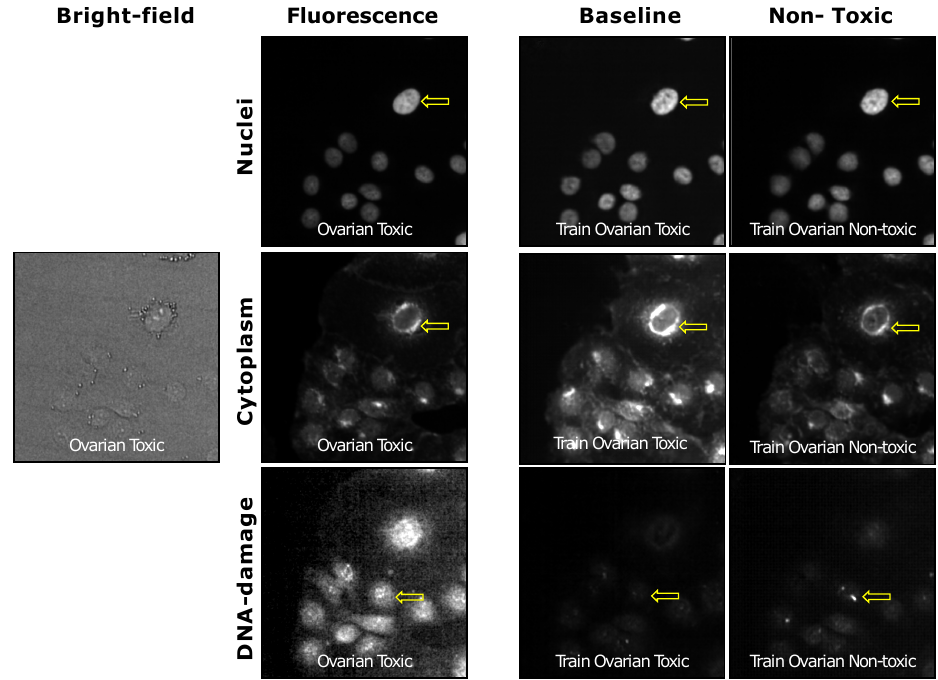}
    \vspace{2mm}
    \caption{\textbf{Qualitative results for the task of generalizing to an unseen phenotype; ovarian toxic from ovarian non-toxic}. Randomly selected bright-field and paired fluorescence for nuclei, cytoplasm and DNA-damage stains alongside the virtual staining predictions from each of the virtual stain models trained on ovarian toxic samples and the virtual stain models trained on ovarian non-toxic samples. We observe the general shape of nuclei and cells are reproduced well relative to the baseline and fluorescence stain. The DNA-damage spots are considerably different from the fluorescence stain for both the baseline and model trained on non-toxic samples. Examples for all three virtual stain tasks are shown by yellow arrows.}
    \label{fig:phenotype_visuals}
\end{figure*}
}

\newcommand\figPhenotypebar{
\begin{figure}[ht]
    \centering
    \includegraphics[width=0.7\linewidth]{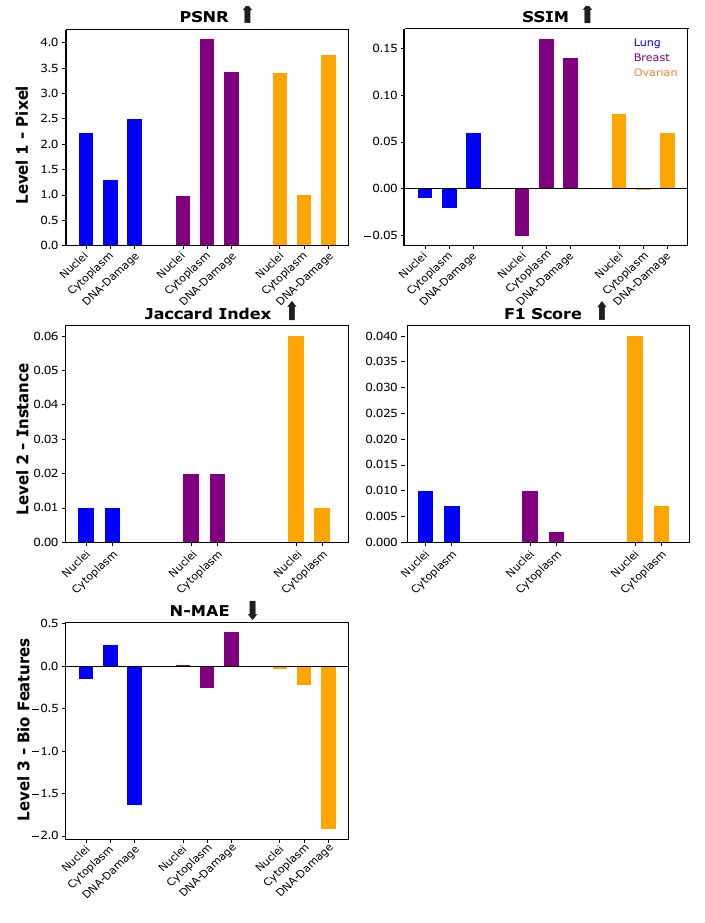}
    \caption{\textbf{Generalization performance of virtual staining models to an unseen phenotype across three levels of evaluation}. Each chart represents the results for that metric, within the chart all virtual stain channels are shown separately and grouped by cell type. Each bar shows the average difference between the virtual stain models trained on non-toxic and the baseline virtual stain models trained on toxic samples. For all three cell types and virtual stain tasks, the PSNR, Jaccard Index, and F1 Score results reveal improved performance from training on non-toxic samples compared to training on toxic samples. Consistently across all metrics, training on ovarian non-toxic leads to improved performance when generalizing to images of ovarian toxic cells.}
    \label{fig:phenotype_bar}
\end{figure}
}

\newcommand\figCelltype{
\begin{figure}[ht]
    \centering
    \includegraphics[width=1\linewidth]{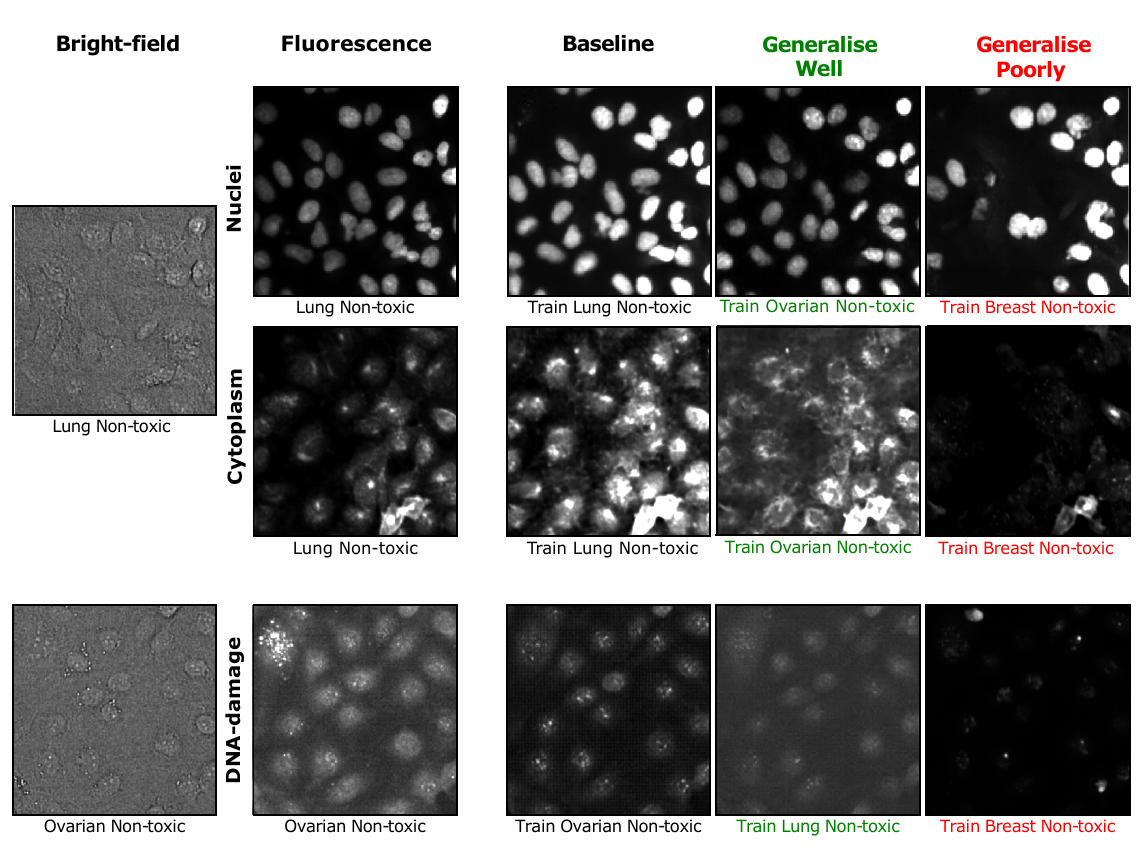}
    \caption{\textbf{Qualitative results for the task of generalizing to unseen cell types}. Randomly selected bright-field and fluorescence for nuclei, cytoplasm and DNA-Damage stains with the virtual stain from the baseline model and the virtual stain models that generalize well and generalize poorly. Both the virtual nuclei and virtual cytoplasm trained on images of ovarian non-toxic cells can reproduce the general shape of the lung cells and in some cases the intensity profile well relative to the baseline model trained on images of lung cells. Meanwhile, the models trained on images of breast cells show a considerable number of nuclei and cytoplasm missing as well as incorrect morphology. For the virtual DNA-damage although the model trained on lung does well relative to the model trained on breast there are still very clear differences in intensity and DNA-damage spot locations between all three virtual stain predictions and the fluorescence stain.}
    \label{fig:celltypefig}
\end{figure}
}

\newcommand\figNMAEov{
\begin{figure*}[ht]
    \centering
    \includegraphics[width=1\linewidth]{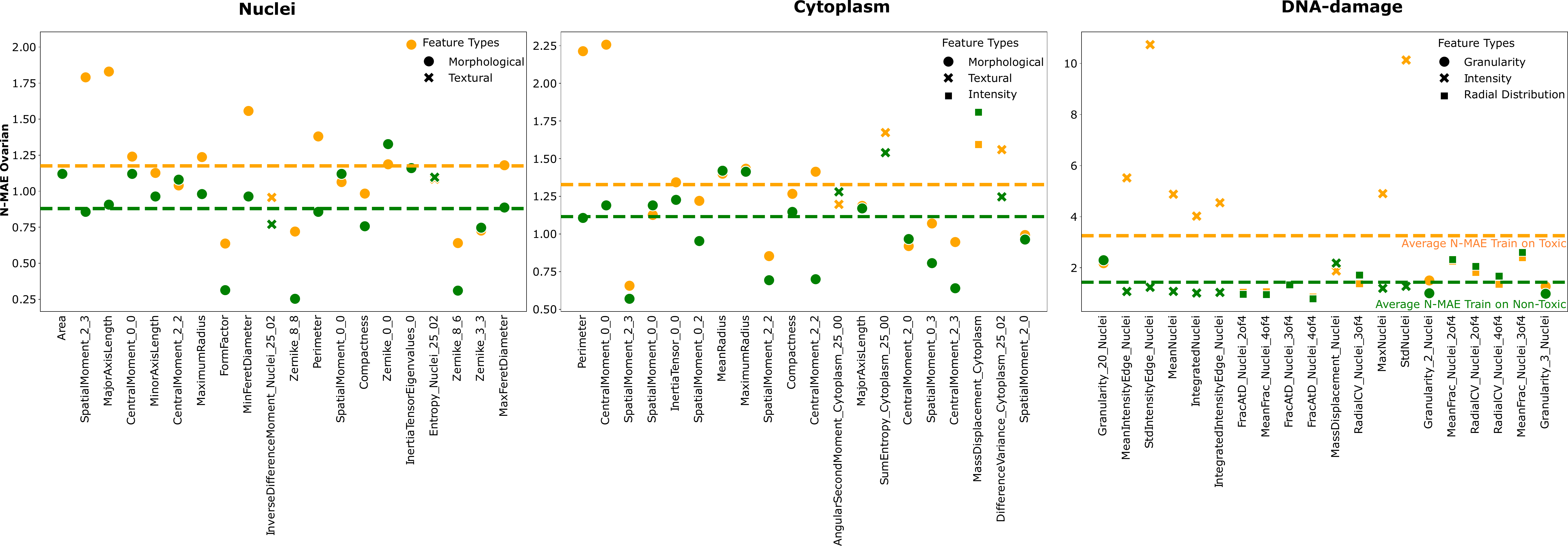}
    \caption{\textbf{N-MAE values of the virtual stain models trained on ovarian non-toxic and ovarian toxic samples for the 20 CellProfiler features identified on the fluorescence stain}. The green line shows the average N-MAE for the virtual stain models trained on ovarian non-toxic samples and the yellow line shows the average N-MAE for the virtual stain models trained on ovarian toxic samples. Across a diverse set of features, training on ovarian non-toxic leads to a biological feature representation that more closely aligns with that found in the fluorescence stains compared to training on ovarian toxic.}
    \label{fig:ov_nmae}
\end{figure*}
}
\newcommand\figheatmapscombined{
\begin{figure}[ht]
    \centering
    \begin{subfigure}[ht]{0.48\textwidth}
        \centering
        \includegraphics[width=\textwidth]{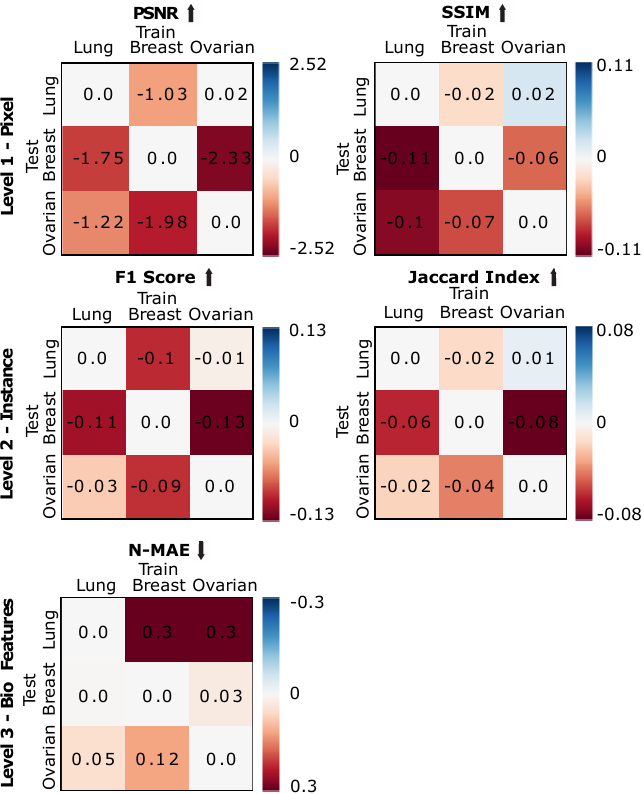}
        \caption{\textbf{Task 2. Generalization to unseen cell types}. The values along the diagonal represent the baseline and are therefore 0. Models trained on breast cell images consistently generalize poorly to other cell types and breast cell images appear hard to generalize too. Levels 1 and 2 show ovarian can generalize well to lung for pixel and instance-level metrics but not when evaluating the biological features. Good generalization to one cell type does not imply the same for other cell types.}
        \label{fig:celltypes}
    \end{subfigure}
    \hfill
    \begin{subfigure}[ht]{0.48\textwidth}
        \centering
        \includegraphics[width=\textwidth]{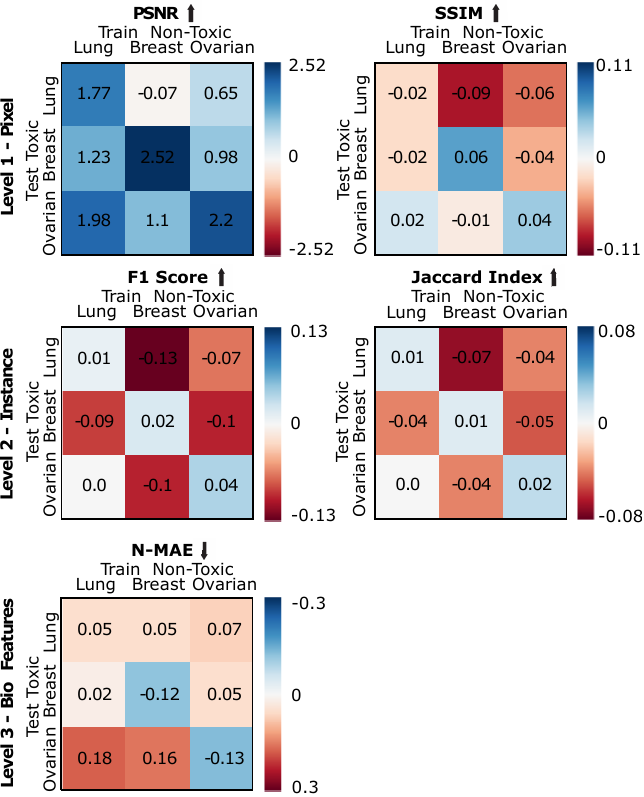}
        \caption{\textbf{Task 3. Generalization to unseen cell types and phenotype}. For each metric, the layout of train/test and the conditional formatting scales are the same as Figure~\ref{fig:side_by_side}a. The values along the diagonal represent the results from generalizing to an unseen phenotype and are kept in for completion. We observe similar patterns of performance when training and testing on images of breast cells to the generalization to unseen cell types as well as considerable improvements in PSNR and N-MAE for several train and test combinations.}
        \label{fig:both}
    \end{subfigure}
    \vspace{2mm}
    \caption{\textbf{Generalization performance of virtual nuclei and virtual cytoplasm models}. Both (a) and (b) have 5 heatmaps with each showing the average results for each metric within the 3 levels of evaluation. Each heatmap shows the cell type train and test set combinations. The metric values are the differences between the models and the baseline model. For each metric, conditional formatting is centered at 0 and scaled to the worst performance across tasks 2 and 3 (to allow for comparison) and its negative, with red indicating bad generalization performance changes and blue positive.}
    \label{fig:side_by_side}
\end{figure}
}

\begin{document}

\maketitle
\begin{abstract}
The large volume and variety of imaging data from high-throughput screening (HTS) in the pharmaceutical industry present an excellent resource for training virtual staining models. 
However, the potential of models trained under one set of experimental conditions to generalize to other conditions remains underexplored. 
This study systematically investigates whether data from three cell types (lung, ovarian, and breast) and two phenotypes (toxic and non-toxic conditions) commonly found in HTS can effectively train virtual staining models to generalize across three typical HTS distribution shifts: unseen phenotypes, unseen cell types, and the combination of both.
Utilizing a dataset of 772,416 paired bright-field, cytoplasm, nuclei, and DNA-damage stain images, we evaluate the generalization capabilities of models across pixel-based, instance-wise, and biological-feature-based levels.
Our findings indicate that training virtual nuclei and cytoplasm models on non-toxic condition samples not only generalizes to toxic condition samples but leads to improved performance across all evaluation levels compared to training on toxic condition samples.
Generalization to unseen cell types shows variability depending on the cell type; models trained on ovarian or lung cell samples often perform well under other conditions, while those trained on breast cell samples consistently show poor generalization. 
Generalization to unseen cell types and phenotypes shows good generalization across all levels of evaluation compared to addressing unseen cell types alone.
This study represents the first large-scale, data-centric analysis of the generalization capability of virtual staining models trained on diverse HTS datasets, providing valuable strategies for experimental training data generation.

\end{abstract}

%%%%%%%%% BODY TEXT
\section{Introduction}
\label{sec:intro}
% \vspace{-2mm}

High-throughput screening (HTS) plays a crucial role in drug discovery by enabling the simultaneous testing of a large number of compounds to assess their effects on cell cultures~\cite{szymanski2011adaptation}.
Fluorescence microscopy is the standard tool in HTS for detecting drug effects on cellular structures~\cite{selinummi2009bright}. By covalently binding different fluorescent dyes to biomolecules (fluorescent staining), it enables biological structures to be simultaneously revealed by the different emission spectra of the dyes, with each dye captured in a separate image channel~\cite{tonks2023evaluation}.

Although it is an essential tool in modern biology, conventional fluorescence microscopy has important practical limitations. The staining protocol typically requires the cells to be fixed and permeabilized - a process in which cells are preserved in their biological state, effectively frozen in time - thus limiting the application of this technique to single time-point studies. 
Furthermore, as the expensive fixation and staining require specialist equipment and the number of fluorescence stains are inherently limited by spectrum saturation, significant interest has been put into label-free microscopy, wherein images of cells are captured without the need for staining~\cite{ounkomol2018label}.
Label-free microscopy~\cite{harrison2023evaluating,gupta2022brightfield}, while cost-effective and scalable, unfortunately, lacks the biological information typically found in the fluorescence stains~\cite{pirone2022stain}.

Recent works have explored the concept of virtual staining to simultaneously leverage the scalability of label-free microscopy and biological information extracted from fluorescence microscopy ~\cite{cross2022label,cross2023class,imboden2023trustworthy,wieslander2021learning,tonks2023evaluation}. 
Virtual staining is typically framed as a multimodal image-to-image translation (I2I) problem~\cite{isola2017image}. In this context, virtual staining models learn to translate unstained microscopy images into the desired labeled images.

While recent works~\cite{cross2022label,cross2023class,imboden2023trustworthy,wieslander2021learning,tonks2023evaluation} have shown the significant potential of virtual staining, the ability of virtual staining models to generalize to images containing variations not present in the training data remains underexplored.
In practice, HTS imaging data is highly diverse, being generated across different imaging systems, experiments, cell types, and phenotypes.
This is known as the generalization gap~\cite{wagner2022make} and has been identified as a key reason for the lack of reusable virtual staining models limiting its potential impact within large-scale applications. 
These challenges are analogous to those in DNA sequence modeling~\cite{avsec2021effective}, where models are typically trained on specific cell types and fail to generalize to new cell types. In offline reinforcement learning~\cite{mediratta2023generalization,cobbe2019quantifying,levine2020offline} where datasets predominantly focus on solving the task in the same environment, limiting the evaluation of generalization to unseen environments.
\figtesaro
Within the context of virtual staining we intend to bridge this generalization gap by performing a systematic data-centric approach to determine whether virtual staining models trained on specific subsets of data can generalize under common distribution shifts. Our approach has two main benefits. First, it would provide guidance on the best data generation practices to produce highly generalizable models. Second, since scientists need to extract biological insights from virtual stains, there must be a framework to quantify the domain of applicability of virtual staining.

In this work, we explore for the first time the generalizability of virtual staining models under three common HTS data distribution shifts: 
\begin{itemize}[label=$\bullet$]
    \item\textbf{Task 1}: Generalizing to new phenotypes
    \item\textbf{Task 2}: Generalizing to new cell types
    \item\textbf{Task 3}: Generalizing to new phenotypes \& cell types
\end{itemize} 

% describe dataset
To investigate these three tasks, we leverage a GSK proprietary dataset of 772,416 images, consisting of bright-field and 3 co-registered widely used fluorescence stains; fluorescein (FITC) for cytoplasm, 6-diamidino-2-phenylindole (DAPI) for nuclei detection and Cyanine (Cy5) for DNA-damage with 3 cell types (ovarian, lung, breast) and 2 phenotypes that correspond to the conditions under which the cells have been cultured  (toxic, non-toxic). We define the wells that contain DMSO (negative control) and the three lowest levels of concentration for 10 GSK prospective compounds to be non-toxic and the three highest concentrations of each compound as well as etoposide and starausporine (positive controls) to be toxic. Example bright-field and fluorescence stain images of non-toxic and toxic sample conditions for each cell type are shown in Figure~\ref{fig:dataset}. 

A total of 54 models were trained; one model for each of the 2 phenotypes $\times$ 3 cell types $\times$ 3 stains $\times$ 3 initializations. A total of 243 individual inference runs were completed (54 for task 1, 54 for task 2 and 135 for task 3) each corresponding to the different train and test combinations. We evaluate outputs using three levels of evaluation; pixel-based, instance-based, and biological-feature-based. As the performance in absolute terms of our method has been reported~\cite{tonks2023evaluation} and because the focus of this work is generalization we report the difference in performance relative to the baseline model trained on the same biological variation found in the training set.
%%%%%%%
%%%%%%%Summary of key findings

For task 1,  we find training only on non-toxic samples generalizes well but can also surpisingly lead to improved performance over training on toxic samples, for virtual nuclei and virtual cytoplasm, even when evaluated on toxic samples. This suggests a new potential strategy for model training that could have broad implications for improving the robustness of virtual staining models when access to phenotype specific data is limited.
% In particular, training on ovarian non-toxic leads to improved performance across virtual nuclei and virtual cytoplasm for all evaluation metrics compared to training on ovarian toxic samples.
For task 2, we find generalizing to unseen cell types to be a mixed result with different generalization outcomes depending on the level of evaluation. These findings underscore the critical challenge of developing models that perform consistently well across diverse conditions.
% Good generalization to one cell type is not always an indicator of good generalization to another cell type. 

For task 3, we observe for virtual nuclei and cytoplasm improvements in pixel-based peak-signal-to-noise ratio (PSNR)~\cite{faragallah2020comprehensive} for the majority of experiments.
Specifically, when trained on ovarian non-toxic, testing on lung toxic and when trained on lung non-toxic test on ovarian toxic we observe good generalization performance across all levels of evaluation.

\indent The performance on task 3, which can be viewed as a combination of tasks 1 and 2, combines the qualitative characteristics of the two simpler tasks. We observe a similar positive effect from training on non-toxic samples as was seen in task 1, and a similarly mixed picture as in task 2 when generalizing to an unseen cell type with particular improvements in PSNR and normalized mean-absolute-error (N-MAE) but a mixture of negative and positive effects on performance across all levels of evaluation.
Consistently across all three tasks, images of breast cells were the hardest to generalize from and generalize to.

Despite good generalization performance across all tasks and several evaluation metrics, virtual DNA-damage predictions are less accurate compared to virtual nuclei and virtual cytoplasm when compared to the fluorescence stain aligning with previous findings~\cite{tonks2023evaluation}.
Overall, we believe this work provides robust insight into potential strategies for generating HTS training data for generalizable virtual staining models. 
% We recommend leveraging non-toxic samples to train more generalizable models for the virtual nuclei and virtual cytoplasm of toxic samples, if you have access to ovarian non-toxic samples we find these to show improved performance across levels of evaluation.
% \vspace{-3mm}
\section{Related Work} 
% \vspace{-2mm}
% Introduce I2I and virtual staining
Virtual staining is a specific formulation of multimodal image-to-image translation (I2I)~\cite{isola2017image}, a machine learning technique in which we want to train a model to translate one modality - label-free brightfield images - to another - fluorescence images. 
Virtual staining using I2I has been widely explored using approaches based on a regression-loss~\cite{ounkomol2018label}, auto-regressive models~\cite{christiansen2018silico}, generative adversarial network (GANs)~\cite{tonks2023evaluation,upadhyay2021uncertainty,cross2022label} and diffusion-based approaches~\cite{cross2023class}. 
Across these methods when tested on samples similar to those in the training sets, virtual staining predictions have been shown to consistently and reliably produce high quality images at the pixel-level~\cite{ounkomol2018label,christiansen2018silico,upadhyay2021uncertainty,cross2022label} that preserve the majority of biological information found in fluorescence images~\cite{tonks2023evaluation,cross2023class}. In order to determine the usability of virtual staining at scale, these systems need to be able to generalize under various data distribution shifts

Despite improvements in virtual staining very few models~\cite{van2021deep,echle2021deep} are known to have been integrated into routine clinical workflows. A recent review~\cite{wagner2022make} of 161 peer-reviewed computational pathology articles identified a core reason being the generalization gap; models failing to maintain performance for unseen data with a shifted distribution. 
Nevertheless, in real-world applications, encountering unseen data is very common.
For example in high-throughput screenings (HTS)~\cite{szymanski2011adaptation} a single imaging machine, among many in a lab, generates imaging data for large numbers of compounds at different concentration levels tested on cell cultures of different cell types (lung or breast) composed of different cell lines (H1299, HCC827). 

Early virtual staining works~\cite{ounkomol2018label} showed models trained on images of hiPSC cells did not perform as well when generalizing to HEK-293, cardiomyocytes and HT-1080 cells. 
Although gross image features were visually comparable, morphological detail improved when the model was trained on data of the same cell type. 
Meanwhile, other virtual staining works~\cite{christiansen2018silico} showed when trained separately on images containing cortical and motor neuron cells generalizing to a single well of a breast cell line (MDE-MD-231), sourced from a new laboratory and new transmitted-light technology led to an increase in pixel-level performance. Similarly, Cross-Zamirski et al.~\cite{cross2023class} trained virtual staining models utilizing a diverse set of 290 cell type and phenotypes from the JUMP Cell Painting Dataset~\cite{Chandrasekaran_Ackerman_Alix_Ando_Arevalo_Bennion_Boisseau_Borowa_Boyd_Brino_et_al._2023} also testing on a single hold out plate, providing limited exploration of performance under domain shift. The question of whether virtual staining models can generalise at scale remains underexplored. 

A trivial solution to generalization would be to train virtual staining models using samples from each domain shift, but in practice training this amount of models is computationally very expensive and not scalable. Alternatively, domain adaptation methods for I2I such as DAI2I~\cite{murez2018image} and 0ST~\cite{luo2020adversarial} have been shown to improve out-of-distribution performance on natural image datasets. None of the aforementioned approaches to bridging the generalization gap have focused on whether certain image domains, when used as training sets, are better able to learn domain-invariant features compared to others. In this paper we 
perform the first large-scale systematic analysis of whether virtual staining models trained on specific subsets of HTS data can generalize under three common distribution shifts.
% \vspace{-3mm}
%--------------------------------
\section{Experiments \& Results}  
%--------------------------------
% \vspace{-2mm}
We first discuss the general points about our dataset, training, inference and evaluation procedures and then the three generalization tasks. 

\indent\textbf{Dataset:} Our experiments are based on a pool of 772,416 individual images generated as part of a GSK HTS study. 
The data comprises 98 16x24 well plates, with each plate containing 384 wells containing a combination of dimethyl sulfoxide (DMSO) (negative control) as it has a relatively low order of systemic toxicity~\cite{pmid:14256595}, 10 GSK candidate compounds with 6 levels of toxicity from low to high and known apoptosis (programmed cell death) inducing compounds etoposide~\cite{pmid:8070030}, and starausporine~\cite{pmid:8913270} with high orders of toxicity (positive controls).
All plates of one cell type have a fixed layout of compounds and controls across the wells. 

We define the wells that contain DMSO and the three lowest levels of toxicity for each compound to be non-toxic and the three highest concentrations of each compound as well as the etoposide and starausporine to be toxic.
Every well consists of 9 fields of view each containing a bright-field and three co-registered fluorescent stains; fluorescein (FITC) for cytoplasm, 6-diamidino-2-phenylindole (DAPI) for nuclei detection and Cyanine (Cy5) for DNA-damage. 
Each cell type was represented by six different cell lines.
\figPhenotypebar
For each cell type and stain, 27,000 bright-field and fluorescence image pairs were sampled from toxic and non-toxic labeled wells separately, with 21,000 image pairs used for training and 6,000 image pairs used for validation.
Random samples for each cell type and labeled wells are shown in Figure \ref {fig:dataset}.
In addition, for each cell type three plates, excluded from any training or validation sets, were used to sample toxic and non-toxic test sets. 
The test sets for both lung and ovarian toxic were 5,562 and for non-toxic were 4,806 while breast toxic was 3,510 and non-toxic was 6,856.
%--------------------------------

\indent\textbf{Training \& Inference}: Using the non-toxic and toxic samples for each cell type, three models were independently trained with different random initialization of weights to translate from bright-field to each fluorescence stain. 
In the following, all performance metrics are computed as the mean performance over all three initializations.
This resulted in a total of 54 models being trained; one model for each of the 2 phenotypes $\times$ 3 cell types $\times$ 3 channels $\times$ 3 initializations. 
This work used the Pix2PixHD ~\cite{wang2018high} architecture with the same hyperparameters as~\cite{tonks2023evaluation}. 
Each model was trained for a maximum of 200 epochs using early stopping. 
A total of 243 individual inference runs were completed each corresponding to the different train and test combinations. 
For task 1, we have 27 trained models ($\times$ 3 cell types $\times$ 3 stains $\times$ 3 initializations) for each model we run inference on the non-toxic and toxic test sets producing 54 in total. For task 2, we have 27 trained models and run inference on the two different cell type test sets producing 54 runs in total. Finally for task 3, we have 9 trained models for each cell type and phenotype (3 stains $\times$ 3 initializations) for the toxic phenotype models we run inference on all three toxic cell type test sets and for the non-toxic only the two toxic test sets with different cell types. This leads to 45 inference runs per cell type totalling 135 plus the 108 from tasks 1 and 2 giving a total 243 runs. Each was evaluated at three levels; pixel-based, instance-based and biological-feature-based. 
%--------------------------------

\indent\textbf{Evaluation: In level 1}, we evaluate performance using two established pixel-level metrics (SSIM~\cite{wang2004image}, PSNR~\cite{faragallah2020comprehensive}). \\
\indent\textbf{In level 2} we evaluate the instance segmentation quality that can be obtained from virtually stained nuclei (DAPI) and cytoplasm (FITC) channels.
We leverage Cellpose~\cite{stringer2021cellpose} to generate instance masks for the fluorescence and virtual staining channels. 
To obtain high-quality segmentations we apply a gamma correction of 0.3 before feeding the images to Cellpose. 
The value of 0.3 was validated by manually checking a subset of random nuclei and cytoplasm fluorescence images and the Cellpose segmentation masks. 
We then compare the masks generated from the fluorescence and virtual staining using common segmentation metrics; Jaccard Index~\cite{jaccard1912distribution} for instance-wise pixel area quality and F1 score~\cite{sasaki2007truth} for object detection.
To compute the F1 score, we use a Jaccard Index~\cite{jaccard1912distribution} threshold of 0.7 in alignment with previous nuclei detection works{~\cite{caicedo2019nucleus}}. \\
\indent\textbf{In level 3}, we utilise a Cellprofiler~\cite{carpenter2006cellprofiler} pipeline that takes the intensity image and Cellpose masks as input to compute instance-wise scores for a collection of 209 morphological, intensity and textural, among other features frequently used in HTS. The types of features are similar to those used in~\cite{tonks2023evaluation} but by extracting these from Cellprofiler~\cite{carpenter2006cellprofiler} we can compute values for each instance instead of a single mean value over all instances in an image
Using a random forest based feature selection method similar to Saabas 2014~\cite{datadive_blog} we extract the twenty most informative features identified on each of the fluorescence test sets and compare the results between the fluorescence channels and those obtained from virtual staining.
The mean absolute error (MAE) between the fluorescence and virtual staining feature scores is computed over each instance, building on previous works~\cite{tonks2023evaluation}. 
To enable the comparison and interpretation of MAE across different feature ranges the resulting MAE for each feature is normalized (N-MAE) by the standard deviation of the fluorescence feature score.
Finally, across all three levels of evaluation, we report the difference between the mean score obtained for each of the trained models and the baseline model trained on images sampled from the same distribution as the test set.
This enables us to compare the change in performance across the different cell type and phenotype combinations relative to each tasks virtual staining baseline.
\figPhenotypevisuals
%--------------------------------
% \vspace{-1mm}

\subsection{Task 1 - Generalization to new phenotype}
% \vspace{-1mm}

We begin by exploring for each cell type, how virtual staining models trained on images containing samples of one phenotype; non-toxic, perform on images containing samples of a different phenotype; toxic. 
We report the difference in performance across the three levels of evaluation between the virtual staining models trained on non-toxic samples and the virtual staining models trained on toxic samples of the same cell type.

\textbf{Training on non-toxic vs toxic improves performance:}
Across all three levels of evaluation, all three cell types, and all three virtual staining tasks shown in Figure~\ref{fig:phenotype_bar}, we find when testing on toxic samples training on non-toxic samples leads to improved results as measured by several metrics compared to training on toxic samples of the same cell type. 
In particular, for all virtual staining tasks we see consistently improved performance in PSNR, Jaccard Index, and F1 score across all cell types as well as the majority of virtual staining tasks showing improved performances in SSIM and N-MAE. 
However, for the task of generalising to lung toxic when training on lung non-toxic for virtual cytoplasm we see consistently improved performance in levels 1 and 2 but worsening results in level 3.
%%%
We believe this is in part due to the heterogeneity of cell expression that occurs in non-toxic healthy cells providing an increase in the diversity of cell states to train generalizable virtual staining models on. In contrast, the toxic cells are induced into specific homogeneous cell states leading to potentially less diversity in training data producing less generalizable models. We can see some evidence of this in Figure~\ref{fig:dataset} where despite non-toxic conditions we observe small amounts of naturally occuring DNA-damage signal in all non-toxic cell types images shown.

\figNMAEov

\textbf{Ovarian non-toxic samples see the largest improvement:}
Across all measurements and virtual staining channels, when generalizing to ovarian toxic samples, training on ovarian non-toxic samples leads to improved performance compared to training on ovarian toxic samples. Upon visual inspection of Figure~\ref{fig:phenotype_visuals} for the virtual nuclei and virtual cytoplasm stains both the baseline and non-toxic models have produced predictions that replicate the general shape and intensity of a large number of cells seen in the fluorescence. 

The prediction of the ovarian virtual nuclei trained on non-toxic is almost indistinguishable from the prediction of the virtual nuclei trained on toxic. However, upon closer inspection, as shown in yellow for certain nuclei the intensity profile and shape of the non-toxic nuclei more closely resembles that found in the fluorescence. Similar findings are highlighted for the cytoplasm intensity profile of individual cells.  
In contrast, the virtual DNA-damage predictions for both trained models are very similar to each other, with the model trained on non-toxic showing a small number of DNA-damage spots but they both display considerable losses of information compared to the fluorescence stain.

\indent Figure~\ref{fig:ov_nmae} shows the N-MAE obtained for the 20 most informative biological features for the virtual staining models trained on ovarian toxic and ovarian non-toxic samples. 
Across all virtual staining tasks, we observe a reduction in the average N-MAE when training on images of ovarian non-toxic. Consistently, we observe this result is not driven by an outlier, but by reductions in N-MAE across a diverse set of biological features.

These findings support the previously shown results that virtual nuclei and cytoplasm models trained on non-toxic samples of specific cell types such as ovarian when generalizing to toxic samples of the same cell type can learn meaningful and diverse biological feature representations that more closely align with the fluorescence stains.
%--------------------------------
% \vspace{-1mm}
\subsection{Task 2 - Generalization to new cell type}
% \vspace{-2mm}
Having explored the generalization to an unseen phenotype, in this section, we focus on the second task of generating virtually stained samples of bright-field images of a different cell type to the images the virtual staining models were trained on.

\figCelltype

\textbf{Generalizing to new cell types is complex:}
Upon visual inspection of Figure~\ref{fig:celltypefig}, we observe that for the virtual nuclei and virtual cytoplasm task the model trained on ovarian cells can reproduce the general shape of the lung nuclei and cytoplasm compared to the baseline model trained on images of lung cells. Although gross image features can be visually seen to be comparable, performance for morphological detail improved when the model was trained on data of the same cell type as also seen in previous works~\cite{christiansen2018silico,ounkomol2018label}. The model trained on breast is unable to reproduce several nuclei and cytoplasm that can be seen in the fluorescence stain.
\figheatmapscombined
Figure~\ref{fig:side_by_side}a shows the different train and test cell type combinations explored for this generalization task and the mean scores for the virtual nuclei and cytoplasm models for each level of evaluation.
% \figCelltypesheatmaps

\indent We observe some high-level patterns of consistently good and bad generalization performance across the different train and test combinations for levels 1 and 2 of our evaluation pipeline.

For all metrics in levels 1 and 2, the virtual nuclei and virtual cytoplasm models trained on images of ovarian cells consistently outperform the corresponding virtual staining models trained on images of breast cells when testing on images of lung cells (shown in row 1) and, in some cases leads to a very small positive effect compared to the baseline model.
% \figgridboth

However, when evaluating the biological features in level 3 we observe a substantial increase in N-MAE relative to the in distribution baseline model, highlighting the importance of evaluating virtual staining in the biological feature space as opposed to using only common pixel-level and instance-level metrics. 
Interestingly, despite virtual models trained on images of ovarian cells performing well on images of lung cells  Figure~\ref{fig:side_by_side}a shows this is not true for the majority of metrics for generalizing to images of breast cells, providing further evidence that the generalization to multiple different cell types is complex and that generalization performance to one cell type does not necessarily mean the same generalization performance to a different cell type. \\
\indent In contrast to the virtual staining models trained on images of ovarian cells, for all three levels of evaluation, the virtual nuclei and cytoplasm models trained on images of breast cells produce scores that show the largest difference in performance compared to the baselines when testing on images of ovarian and lung cells. 

\indent We observe that models trained on images of ovarian or lung cells do not generalize well to images of breast cells.
This suggests that images of breast cells are not as good as lung or ovarian for training virtual staining models that generalize well to unseen cell types and they are also difficult to generalize to.
Finally, for the virtual nuclei and virtual cytoplasm results, we observe a lack of symmetry in performance across the different cell types for each of the measurements.
The large majority of results do not show that performance when training on one cell type and testing on another is very similar if the train and test cell types are interchanged.

For the virtual DNA-damage models (not shown in Figure~\ref{fig:side_by_side}a), similar to virtual nuclei and cytoplasm, training on images of breast cells and testing on images of ovarian and lung cells consistently leads to bad generalization across all metrics.
Meanwhile, training on images of lung cells when testing on images of ovarian and breast cells leads to small positive effects across PSNR and SSIM as well as a very small increase in N-MAE. \\

Examining the individual test set image scores for each cell type generalization task across all three virtual staining models, for all metrics we find that the differences are consistent across multiple images and not driven by a small number of outliers.

%--------------------------------

% \vspace{-1mm}
\subsection{Task 3 - Generalization to new phenotypes \&  new cell types}
% \vspace{-2mm}
In the final section, we combine the two previous distribution shifts and explore how virtual staining models trained on images of cells in non-toxic conditions of one cell type generalize to cells of a different cell type in toxic conditions. 

We report the difference in performance across the three levels of evaluation between the virtual staining models trained on non-toxic samples and the virtual staining models trained on toxic samples of the same cell type.

\textbf{Generally good generalization performance}
In a similar approach to 
Figure~\ref{fig:side_by_side}a, Figure~\ref{fig:side_by_side}b shows the different non-toxic training and toxic test cell type image set combinations explored for this generalization task and the corresponding mean scores for the virtual nuclei and virtual cytoplasm models for each of the three levels of evaluation. 

We observe similar differences between the performance results within levels 1 and 2 against level 3 found in task 1 when training on non-toxic lung cells and testing on toxic ovarian showing improvements in pixel-wise and instance-wise metrics relative to the baseline but increases in N-MAE when evaluating the biological feature representation.

In general, we see considerable improvements across all metrics compared to those shown in task 2 (See Figure~\ref{fig:side_by_side}a)  which supports our findings from the phenotype generalization task.
In particular, we observe an increase in positive PSNR results and a reduction in the highest N-MAE from 0.3 to 0.18. 
However, in some cases, such as when testing on ovarian toxic we observe an increase in N-MAE.
For levels 1 and 2 we observe similar reduced generalization performance when training and testing on images of breast cells.

%--------------------------------
% \vspace{-3mm}
\section{Discussion \& Conclusion}
% \vspace{-1mm}
We have investigated the generalization performance of virtual nuclei, virtual cytoplasm and virtual DNA-damage models for three common HTS generalization tasks. 
Firstly, generalizing to an unseen phenotype. 
Secondly, generalizing to unseen cell types. 
Finally, generalizing to an unseen phenotype and cell types combined.
Performance has been evaluated using metrics at the pixel level, instance level and biological feature level. 

% Phenotype
For the first generalization task, for both virtual nuclei and virtual cytoplasm, training on non-toxic samples leads to both good generalization and improved performance relative to training on toxic samples across all levels of evaluation.
In particular, when testing on toxic ovarian samples, training on non-toxic ovarian samples produced predictions that achieved higher pixel-level quality, more accurate instance-wise translation and scores for the majority of biological features that more closely match the scores found in the fluorescence channels compared to training on toxic ovarian samples.

Previous work on the generation of The JUMP Cell Painting Dataset~\cite{Chandrasekaran_Ackerman_Alix_Ando_Arevalo_Bennion_Boisseau_Borowa_Boyd_Brino_et_al._2023} suggests it is necessary to have training sets with a high diversity of phenotypes to effectively train machine learning models. However, these results demonstrate that for this specific HTS dataset and the explored virtual staining tasks, training on non-toxic samples alone can lead to both good generalization and also performance improvements when measured using a variety of evaluation metrics. 
As such, we believe that for virtual nuclei and virtual cytoplasm staining of toxic samples training on widely available non-toxic samples is a viable alternative to training on toxic samples. 
These findings could lead to a reduction in the number of cell- and phenotype-specific models needed to efficiently utilize virtual staining for diverse HTS datasets. 

% Cell type
For the second generalization task, we find generalizing to unseen cell types is complex,  we identify for pixel-wise and instance-based metrics, training on ovarian and testing on lung led to good generalization performance for the virtual nuclei and virtual cytoplasm. 
However, when evaluating at the biological feature level we observed relatively high differences in N-MAE values.
These results reveal that determining the correct evaluation metric for a chosen virtual staining application is important to effectively evaluate whether certain cell type-specific training sets can better generalize compared to other cell types.
% In the case where instance segmentation is the ultimate objective prioritize images of non-toxic ovarian cells, if the intended use of the virtual stained outputs for feature extraction then a training set of non-toxic lung cells should be prioritized. 
Additionally, we find that good generalization to one unseen cell type does not necessarily mean the same can be expected for another cell type. In contrast, we observe that training on images of breast cells consistently leads to bad generalization performance across all levels of evaluation.
We also found that models trained on other cell types were poor at generalizing to images of breast cells.
Further analysis into the three non-toxic cell type data sets revealed after a manual inspection of a random subset of 5,000 images from each cell type that breast cells are distributed more sparsely than the other cell types (Figure~\ref{fig:dataset}).
The average number of cells in the inspected breast images was less than half of that in ovarian and a third of that in lung. 
This corresponds to the anatomical function of breast cells, which require more fat in their surroundings~\cite {ellis2013anatomy}, reducing the number of cells in any image as shown in Figure~\ref{fig:dataset}.

\indent On the other hand endocrine signaling~\cite{Cooper_GM} of ovarian cells requires they be closely packed together, and the vital role of lung cells to exchange gas, in principle requires more cells. 
We believe that these structural differences are responsible for the poor generalization performance of images of breast cells.
An additional possible explanation could be that the lower density of breast cells corresponds to a smaller volume of training data for the virtual staining models.

For the final generalization task, we see the same challenges when training on and generalizing to images of breast cells for the virtual nuclei and virtual cytoplasm. However, in general, across all levels of evaluation and the explored celltype and phenotype combinations we find that training on non-toxic is preferable even when additionally generalizing to toxic samples of an unseen cell type (See Figure~\ref{fig:side_by_side}b compared to Figure~\ref{fig:side_by_side}a).
We saw particular improvements in PSNR and N-MAE and relatively similar performance for SSIM, F1 Score and Jaccard Index.

\indent Despite what appears to be good virtual DNA-damage generalization performance across certain evaluation metrics for all three generalization tasks, when examining the absolute values the results are poor compared to virtual nuclei and virtual cytoplasm affirming the findings from previous work~\cite{tonks2023evaluation}.
This becomes clear qualitatively in Figure~\ref{fig:phenotype_visuals} and Figure~\ref{fig:celltypefig} and quantitatively in Figure~\ref{fig:ov_nmae} where the N-MAE values are noticeably larger than in the other two channels.

Future work should explore the generalization performance to a broader selection of unseen cell types, and investigate further the issues with producing accurate virtual DNA-damage stains.
\newpage

\bibliographystyle{plainnat}
\bibliography{references}
\end{document}